\documentclass[letterpaper]{article} 
\usepackage{aaai2026}
\usepackage{times}  
\usepackage{helvet}  
\usepackage{courier}  
\usepackage[hyphens]{url}  
\usepackage{graphicx} 
\usepackage{natbib}  
\usepackage{caption} 
\usepackage{algorithm}
\usepackage{algorithmic}

\usepackage{booktabs}
\usepackage{amsmath}
\usepackage{amssymb}
\usepackage{enumitem}

\usepackage{newfloat}
\usepackage{listings}
\DeclareCaptionStyle{ruled}{labelfont=normalfont,labelsep=colon,strut=off} 
\floatstyle{ruled}
\newfloat{listing}{tb}{lst}{}
\floatname{listing}{Listing}
\title{Switch-Hurdle: A MoE Encoder with AR Hurdle Decoder for Intermittent Demand Forecasting}
\author{
    Fabian Mu\c{s}at,\\
    Simona C\u{a}buz,
}

\title{Switch-Hurdle: A MoE Encoder with AR Hurdle Decoder for Intermittent Demand Forecasting}
\author {
    Fabian Mușat, 
    Simona Căbuz
}
\affiliations{
    eMAG, Romania\\
    fabian.musat@emag.ro, simona.cabuz@emag.ro
}

\usepackage{bibentry}

\begin{document}

\maketitle

\begin{abstract}
%
%

Intermittent demand, a pattern characterized by long sequences of zero sales punctuated by sporadic, non-zero values, poses a persistent challenge in retail and supply chain forecasting.
Both traditional methods, such as ARIMA, exponential smoothing, or Croston variants, as well as modern neural architectures such as DeepAR and Transformer-based models often underperform on such data, as they treat demand as a single continuous process or become computationally expensive when scaled across many sparse series.
To address these limitations, we introduce Switch-Hurdle: a new framework that integrates a Mixture-of-Experts (MoE) encoder with a Hurdle-based probabilistic decoder.
The encoder uses a sparse Top-1 expert routing during the forward pass yet approximately dense in the backward pass via a straight-through estimator (STE). 
The decoder follows a cross-attention autoregressive design with a shared hurdle head that explicitly separates the forecasting task into two components: a binary classification component estimating the probability of a sale, and a conditional regression component, predicting the quantity given a sale.
This structured separation enables the model to capture both occurrence and magnitude processes inherent to intermittent demand.
Empirical results on the M5 benchmark and a large proprietary retail dataset show that Switch-Hurdle achieves state-of-the-art prediction performance while maintaining scalability.
\end{abstract}


\section{Introduction}
\label{sec:introduction}
Accurate prediction of retail demand directly impacts inventory optimization, warehousing efficiency and capital utilization.
Inaccurate results, particularly in high-volume, high-SKU environments lead to significant financial penalties, ranging from increased rental costs due to overstocking to reduced customer service levels caused by stock-outs.
The most difficult challenge in this domain stems from intermittent demand, a common pattern across retail, characterized by sporadic bursts of sales combined with long periods of zero activity \cite{garza2023timegpt}.
Both traditional forecasting methods, such as ARIMA or exponential smoothing, as well as contemporary state-of-art deep learning methods, like DeepAR or Transformer variants, often struggle with this dual-state data. 
%
These models typically treat the entire time series as single processes, leading to biased architectures towards smoothing.

Modern Transformer based approaches leverage attention and encoder-decoder structures to capture long-range dependencies in time series patches.
However, scaling these solutions is computationally expensive and most often they still fail to capture the fundamental heterogeneity of intermittent data.

To overcome these limitations we introduce Switch-Hurdle: a new Mixture-of-Experts (MoE) encoder decoder architecture designed to address probabilistic forecast of intermittent demand.
%
%
Our contributions are as follows:
\begin{itemize}
    \item Switch–Hurdle architecture: A Transformer with an approximately dense expert routing in the encoder and a lightweight cross-attention autoregressive decoder tailored for intermittent demand.
    \item Hurdle head for zeros and deficits: A shared hurdle head that separates “whether a sale happens” from “how much is sold,” handling both zero-inflation and zero-deflation. 
    \item Covariate integration and training dynamics: A simple AR decoding scheme that combines prior predictions with future covariates, alongside a stable training recipe supporting probabilistic and point-wise objectives.
    \item Empirical gains: State-of-the-art WRMSSE on M5 and improved WAPE on an internal dataset, supported by ablations and expert-routing analyses.
\end{itemize}

The remainder of the paper is organized as follows: Section~\ref{sec:related-work} reviews related work.
Section~\ref{sec:method} presents the proposed Switch-Hurdle architecture, detailing the Switch MoE encoder, cross-attention autoregressive decoder, hurdle head, and training objectives.
Section~\ref{sec:experiments} describes datasets, metrics, and the experimental setup, reports results on M5 and the internal dataset, including baselines and ablations and provides a conditional expert routing analysis.
Section~\ref{sec:conclusion} concludes with limitations and future directions.

\section{Related Work}
\label{sec:related-work}
While tree-based ensembles have long dominated time-series forecasting, Transformer-based methods have recently gained prominence. Contemporary approaches fall into three groups: (i) Transformer-based models, (ii) Mixture-of-Experts (MoE) architectures, and (iii) large Foundation Models (FMs).

\subsubsection{Transformers for time series forecasting}
The shift from recurrent and convolutional models to Transformers improved the modeling of both short and long range dependencies via self-attention. 
On M5 and related benchmarks, vanilla Transformer, Informer, and TFT outperform AutoARIMA and AutoETS, with MASE gains of about 26–29\% and WQL reductions \cite{caetano2025transformer,jintime}, albeit at higher computational cost \cite{oliveira2024evaluating}. 
Efficiency oriented variants target attention complexity or tokenization: Autoformer introduces sparse mechanisms \cite{wu2021autoformer}; PatchTST uses patch tokenization and improves robustness and accuracy \cite{nietime}; LipFormer removes costly components, like LayerNorm or positional encodings, and integrates future covariates via a Dual Encoder, yielding gains across backbones \cite{wang2025towards,en18185000}.

\subsubsection{Mixture of Experts for time series forecasting}
MoE scales the model capacity by activating a sparse subset of experts per token. 
Switch Transformers popularized Top-1 token-wise routing \cite{fedus2022switch}. 
Subsequent work explored soft or continuous mixtures and improved balancing \cite{puigcerversparse}, and Dense-to-Sparse training that begins dense and gradually sparsifies routing \cite{nie2022densetosparse}. 
%

\subsubsection{Foundation Models for time series forecasting}
Large FMs pre-trained on broad time-series corpora demonstrate zero-shot generalization. 
MOMENT \cite{goswami2024moment}, which is trained on the Time-series Pile, uses a patch-style tokenization similar to PatchTST. 
LLM-based approaches adapt prompting for time series \cite{jia2024gpt4mts}, while domain specific time-series FMs scaled via MoE (e.g., Time-MoE \cite{shitime}, Moirai MoE \cite{liumoirai}) reach billions of parameters and strong benchmark performance. 

Despite recent advancements, in real retail settings, a gap remains: even though Transformers beat statistical baselines, they often lag behind feature-rich methods like LightGBM and DeepAR on the hierarchical WRMSSE metric, which favors good aggregation and strong use of external features. 
Moreover, often times, dense models can bias toward smoothing, which is problematic for sparse, zero-inflated retail demand.
This points to the need for models that better use covariates and explicitly handle zero inflation.

Compared to prior work, we pair a lightweight cross-attention AR decoder with a hurdle head to explicitly model zero occurrence and positive-demand magnitude, while using MoE routing to specialize representations.

\section{Method}
\label{sec:method}
\subsection{Model Architecture}

\begin{figure*}[t]
\centering
  \includegraphics[width=0.7\textwidth]{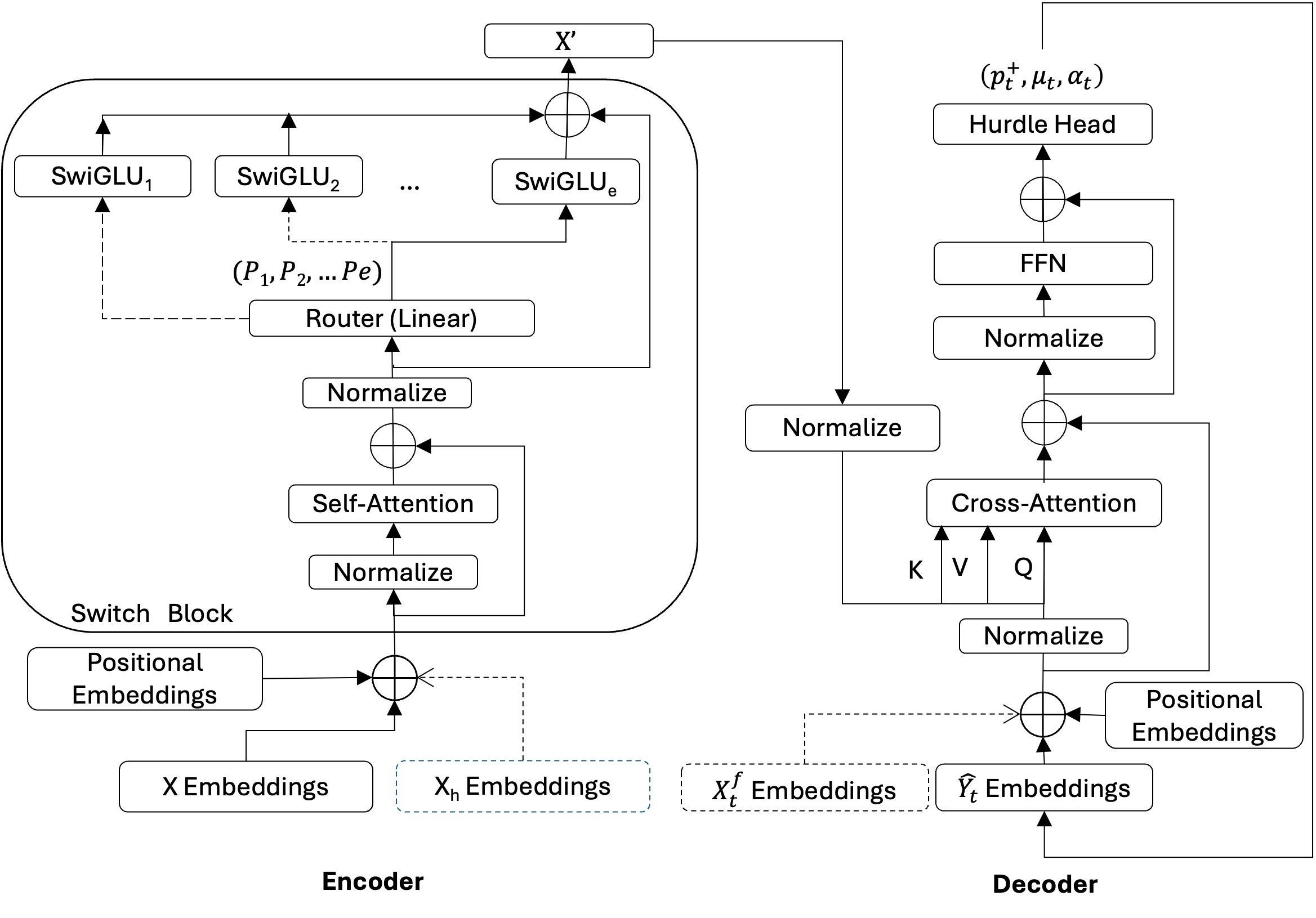}
    \caption{
    \textit{Main architecture for the Switch-Hurdle Transformer.}
    The encoder (left) uses Top-1 MoE routing with SwiGLU experts to extract specialized representations of demand and covariate embeddings.
The decoder (right) applies cross-attention over the encoder’s context memory to generate step-wise probabilistic forecasts $(p_t^{+}, \mu_t, \alpha_t)$.
Each step conditions on the previous prediction, future covariates, and positional embeddings, while the shared hurdle head jointly models zero-demand probability and the conditional Negative-Binomial distribution for positive demand.}
  \label{high_level}
\end{figure*}

As depicted in Figure~\ref{high_level}, our solution combines a dense-to-sparse MoE encoder with a lightweight autoregressive (AR) decoder and a shared hurdle head.
This design addresses two key challenges in intermittent demand forecasting: (i) sparsity and zero-inflation, and (ii) the need for multi-step forecasts.
The encoder captures temporal and categorical structure through a two-layer MoE; the decoder models the conditional distribution of future demand via a probabilistic hurdle mechanism.

%
%


\subsection{The Switch MoE Encoder} 

Inspired from the Switch Transformer's work, \cite{fedus2022switch} our Switch Block works as a pattern recognition component, routing the essential information to the decoder. This mechanism ensures experts' unsupervised specialization, partitioning the token space into clusters of dedicated patterns. 

The encoder essentially routes each token representation through a Mixture-of-Experts (MoE) layer composed of $E$ SwiGLU experts $f_e(\cdot)$.
%
%
%
%
Each expert is a position-wise feed-forward network parametrized by $\{W_{1,e}, W_{2,e}, W_{3,e}\}$ that applies a gated activation using SwiGLU, i.e.
\begin{equation}
f_e(x) \;=\; (\mathrm{Swish}(xW_{1,e}) \odot (xW_{2,e})\big)W_{3,e},
\end{equation}
where $\mathrm{Swish}(\cdot)$ is the Swish function with $\beta = 1$, i.e. Sigmoid Linear Unit (SiLU).

Now let $x_t \in \mathbb{R}^{d}$ denote the per-token embedding entering the MoE, with $t \in [1 \dots L]$.
A learned router $W_g \in \mathbb{R}^{E \times d}$ produces logits $r_t = W_g x_t$ and routing probabilities $p_t=\mathrm{softmax}(r_t)$.

To encourage expert specialization and improve prediction performance, we adopt a Top-1 straight-through (STE) gate: 
the forward pass makes a hard, one-hot expert choice per token, while the backward pass uses the soft probabilities to provide stable gradients. 
At inference, we evaluate only the selected expert per token.
This discrete routing concentrates capacity into distinct experts rather than diluting it in a single dense FFN. A secondary benefit is practical efficiency at inference time, where only the selected expert is evaluated.
Given the routing probabilities $p_t \in \mathbb{R}^{E}$, we obtain:
\begin{equation}
g_t \;=\; \mathrm{one\_hot}\!\big(\arg\max(p_t)\big) \;+\; \big(p_t - \mathrm{stopgrad}(p_t)\big).
\end{equation}
Here, $\mathrm{stopgrad}(\cdot)$ returns its argument in the forward pass but blocks gradients, making the gate forward-sparse (Top-1) and backward-dense.

The MoE output for step $t$ is
\begin{equation}
x'_t \;=\; \sum_{e=1}^{E} g_{t,e}\, f_e(x_t) \;\in\; \mathbb{R}^{d}.
\end{equation}
Stacking over the context yields the encoder memory
\begin{equation}
X' \;=\; \big[x'_1,\ldots,x'_L\big]^\top \;\in\; \mathbb{R}^{L \times d},
\end{equation}
which serves as the decoder’s cross-attention memory.

To ensure balanced expert utilization, we regularize the pre-gate probabilities with a KL-to-uniform term:
\begin{align}
\bar p_e &= \frac{1}{N}\sum_{i=1}^{N} p_{i,e}, \\
\mathcal{L}_{\text{balance}} &= \sum_{e=1}^{E} \bar p_e \log\!\frac{\bar p_e}{1/E},
\label{eq:L_balance}
\end{align}
where $N$ is the number of tokens in the minibatch (time steps $\times$ series).
This replaces the Switch auxiliary loss and promotes stable specialization without collapse.

\begin{figure}[t]
\centering
\includegraphics[width=\linewidth]{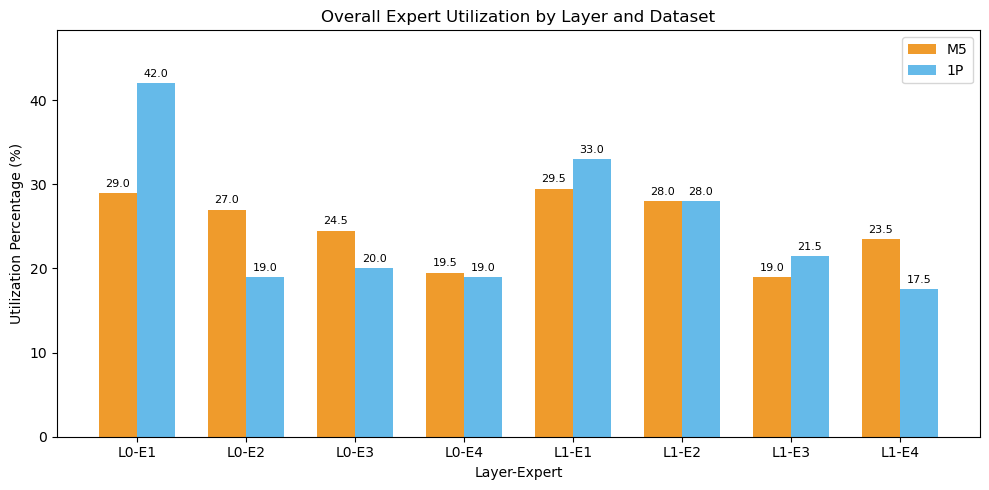}
\caption{
\textit{Overall expert utilization by layer and dataset.}
Bars show the percentage of tokens routed to each expert for the two Switch-encoder layers (L0, L1) on M5 and internal 1P data after introducing the KL-to-uniform regularizer.
The distribution is balanced without collapse while still reflecting dataset- and layer-specific specialization.
}
\label{fig:expert_utilization}
\end{figure}

Figure~\ref{fig:expert_utilization} illustrates balanced routing induced by the KL regularizer, while the Top-1 STE preserves specialization across layers and datasets.
This supports our design choices: Top-1 straight-through routing provides sparse forward selection with dense gradient updates, and the KL-to-uniform regularizer prevents collapse while allowing specialization.

\subsection{The Autoregressive Hurdle Decoder}
\label{sec:ar-decoder}

The decoder is a cross-attention only block that produces hidden states $h_t$ for $t=(1,\ldots,T)$.
Each $h_t$ is subsequently mapped to the forecast parameters $(p_t^{+}, \mu_t, \alpha_t)$ by the hurdle head, defined in Sec.~\ref{sec:hurdle-head}.
%
%
%
At step $t$, the decoder conditions on (i) the previously predicted demand $\hat{y}_{t-1}$, (ii) the step-specific future covariates $X^{f}_t$ (e.g., promotions or planned price changes), and (iii) a learned positional embedding $p^{(\mathrm{dec})}_t$:
\begin{equation}
q_t \;=\; W_p\,\log\!\big(1+\hat{y}_{t-1}\big) \;+\; W_f\,X^{f}_t \;+\; p^{(\mathrm{dec})}_t,
\end{equation}
where $W_p$ and $W_f$ are learned projections.

We use $\log(1+\hat{y}_{t-1})$ to stabilize scale and gradients for count data since it handles zeros and downweights large spikes better; other monotone links such as $\log(\varepsilon+\cdot)$ or a learned embedding are also viable.

Each $q_t$ attends to the encoder output $X' \in \mathbb{R}^{L \times d}$ via cross-attention, yielding a hidden state $h_t$ that integrates historical context with planned future signals.
The decoder has no self-attention.
Autoregression arises solely from the inclusion of $\hat{y}_{t-1}$ in $q_t$.
We apply teacher forcing with a scheduled ratio, gradually replacing ground-truth $y_{t-1}$ with model predictions $\hat{y}_{t-1}$ during training.%

%
%
The reason we adopt an AR decoder, conditioning on $\hat{y}_{t-1}$ is to (i) reduce exposure bias while keeping inference simple and fast, and (ii) avoid redundant lag inputs, since long-range information is already available through cross-attention to $X'$.
%

\subsection{Hurdle Head for Demand Distribution}
\label{sec:hurdle-head}
The decoder output $h_t$ is passed to a shared hurdle head that jointly predicts:
(i) the probability of a positive demand event $p_t^{+}$, and
(ii) the parameters $(\mu_t,\alpha_t)$ of a conditional Negative-Binomial (NB) distribution for positive counts:
\begin{align}
p_t^{+} &= \sigma(w_p^\top h_t), \\
\mu_t   &= \mathrm{softplus}(w_\mu^\top h_t + b_\mu), \\
\alpha_t&= \mathrm{softplus}(w_\alpha^\top h_t + b_\alpha).
\end{align}
The NB probability of zero is:
\begin{align}
p_{0,t} = \left(1+\alpha_t \mu_t\right)^{-1/\alpha_t}.
\end{align}
The resulting hurdle distribution is
\begin{align}
P(Y_t=0) &= 1 - p_t^{+}, \\
P(Y_t=y>0) &= \frac{p_t^{+}}{1-p_{0,t}} \; \\
P_{\mathrm{NB}}(Y_t&=y;\mu_t,\alpha_t),
\end{align}
where $P_{\mathrm{NB}}$ denotes the NB pmf under the mean–dispersion parameterization:
\[
P_{\mathrm{NB}}(y;\mu,\alpha)
= \frac{\Gamma(y+\tfrac{1}{\alpha})}{\Gamma(\tfrac{1}{\alpha})\,y!}
\left(\frac{1}{1+\alpha\mu}\right)^{\tfrac{1}{\alpha}}
\left(\frac{\alpha\mu}{1+\alpha\mu}\right)^{y}.
\]
This formulation separates the zero-occurrence process from positive-demand magnitude, yielding calibrated probabilistic forecasts for intermittent series.

We use NB for the positive-count component because it models overdispersion, $\mathrm{Var}=\mu_t+\alpha_t \mu_t^2$, common in retail demand, while Poisson forces $\mathrm{Var} =\mu $ and underestimates uncertainty.
%
The hurdle head is modular and can be swapped with any count distribution.

\subsection{Training Objectives}
The Switch-Hurdle model is optimized with one of two composite objectives, both including the load-balancing term $\mathcal{L}_{\text{balance}}$ in Eq.~\eqref{eq:L_balance}.

\paragraph{Probabilistic Objective.}
For distributional calibration, we optimize the hurdle negative log-likelihood (NLL) with load balancing:
\begin{equation}
\mathcal{L}_{\text{Prob}} = \mathcal{L}_{\text{Hurdle}} + \lambda_{\text{aux}}\,\mathcal{L}_{\text{balance}}.
\label{eq:probabilistic_objective}
\end{equation}

\paragraph{Point-wise Hybrid Objective.}
For deterministic accuracy, we combine MAE with the probabilistic structure and anneal its weight:
\begin{equation}
\mathcal{L}_{\text{Total}} = \mathcal{L}_{\text{MAE}} + \lambda_{\text{decay}}\big(\mathcal{L}_{\text{Hurdle}} + \mathcal{L}_{\text{balance}}\big),
\label{eq:total_loss}
\end{equation}
where $\lambda_{\text{decay}}$ is initialized to $1.0$ and reduced by $30\%$ after each epoch until a floor of $0.05$, allowing the model to focus increasingly on MAE in later epochs.

\section{Experiments}
\label{sec:experiments}

We evaluated our variant of the Switch-Hurdle Transformer model on proprietary, internal datasets and on the external, popular, M5 Walmart dataset (Makridakis et al., 2022). Both tasks aim to forecast demand at the daily level per SKU, with a horizon of a month. M5 contains $30,490$ time series with static features such as store, item and state id, and time-varying features such as past events, campaigns, etc. 
Our internal dataset contains about $40,000$ time series with static features such as product and vendor id and time-varying features such as past events, promos, campaigns, price-related features, etc. The complexity of both datasets is similar, with M5 being more challenging to predict due to the smaller feature range and variable demand types and peaks. We use a horizon of $T = 28$ and a context or input length of $L = 56$. 

For both datasets, we tested two types of training: 
\begin{itemize}
    \item probabilistic: we optimize the loss of the hurdle model, made up of a binary cross-entropy part, for the Bernoulli process, combined with a negative log-likelihood of a truncated negative binomial distribution, for the positive demand part;
    \item point-wise: we optimize the loss of the hurdle model and a point loss such as mean absolute error (MAE).
\end{itemize}

\subsection{Base results}

For our internal datasets, we first took a small (10\%) sample that uniformly represents our data, so we can draw some quick conclusions. We started by measuring basic models to establish a baseline.

\begin{table}[ht]
\centering
\caption{
\textit{Classical baseline metrics on the internal sample dataset.}
This table compares traditional forecasting methods on a 10\% stratified sample of our internal data.  
Results show that the Croston model, specifically designed for intermittent demand, achieves the best overall accuracy, followed by SARIMAX.  
These results confirm the zero-inflated nature of the dataset and motivate the use of models that explicitly separate demand occurrence and magnitude, such as the proposed hurdle-based approach.
}
\begin{tabular}{|l || c | c|}
\hline
\textbf{Model} \rule{0pt}{1em} & \textbf{WAPE} & \textbf{MASE}  \\
\hline
Naïve  \rule{0pt}{1em} & 111.14\% & 1.2064 \\
SARIMAX (1,1,0)    & 101.71\% & 1.0757 \\
SARIMAX (1,1,1)    & \underline{97.67\%} & \underline{1.0330} \\
Croston (Biased)   & \textbf{93.15\%} & \textbf{0.9852} \\
\hline
\end{tabular}
\label{tab:sample_classical}
\end{table}

Unsurprisingly, Croston - which is very well suited for intermittend demand - has the best metrics, with SARIMAX coming in a close second.

\subsection{Sample results}

 On this small (10\%) sample, we trained some well-established deep learning models using both methods described above and chose the top Transformer-based model to put it up against our Switch-Hurdle Transformer variant.

 We also compare our results against zero-shot Time-MoE, a recent time-series foundation model with a Mixture-of-Experts architecture, on our internal data. 
 We evaluate its point forecasts using WAPE and MASE.

\begin{table*}[!htbp]
\centering
\begin{minipage}[t]{0.49\textwidth}
\centering
\caption{\textit{Probabilistic (NLL) Objective.} 
}
\label{tab:sample_tests_nll}
\begin{tabular}{|l || c | c|}
\hline
\textbf{Model} \rule{0pt}{1em} & \textbf{WAPE} & \textbf{MASE}  \\
\hline
PatchTST  \rule{0pt}{1em} & 121.13\% & 1.2976 \\
TFT              & \textit{80.23\%} & \textit{0.8595} \\
DeepAR           & \underline{79.27\%} & \underline{0.8492} \\
Switch-Hurdle (ours) & \textbf{74.70\%} & \textbf{0.8108} \\
\hline
\end{tabular}
\end{minipage}
\hfill
\begin{minipage}[t]{0.49\textwidth}
\centering
\caption{\textit{Point-Wise (MAE) Objective.} 
}
\label{tab:sample_tests_mae}
\begin{tabular}{|l || c | c|}
\hline
\textbf{Model} \rule{0pt}{1em} & \textbf{WAPE} & \textbf{MASE}  \\
\hline
PatchTST  \rule{0pt}{1em} & 92.82\% & 0.9944 \\
DeepAR           & 66.39\% & 0.7112 \\
Time-MoE (Zero-Shot)$^\dagger$             & 84.77\% & 1.0438 \\
TFT              & \underline{61.92\%} & \underline{0.6634} \\
Switch-Hurdle (ours) & \textbf{56.97\%} & \textbf{0.6184} \\
\hline
\end{tabular}
\end{minipage}
\caption*{Deep learning benchmark metrics on the sample dataset. 
\textbf{Left Table} (Probabilistic Objective): Models trained with the Hurdle Loss ($\mathcal{L}_{\text{Hurdle}}$) and load balancing ($\mathcal{L}_{\text{balance}}$). 
\textbf{Right Table} (Point-Wise Objective): Models trained with the hybrid loss ($\mathcal{L}_{\text{Total}}$), optimized for deterministic accuracy.\\
\footnotesize{$^\dagger$Time-MoE results are obtained by applying the released Time-MoE model on our internal dataset and converting its probabilistic output to point forecasts via the predictive mean.}
}
\end{table*}

\subsubsection*{Notes}
\label{sample:notes}
\begin{itemize}
  \item \textbf{Note 1\label{note:switch_nll}} - Switch-Hurdle Transformer was trained using a combination of binary cross-entropy and truncated negative binomial negative log-likelihood as the hurdle model imposes it and auxiliary loss (load balancing).
  \item \textbf{Note 2\label{note:switch_mae}} - Switch-Hurdle Transformer was trained using a combination of mean absolute error and auxiliary loss (load balancing).
  \item \textbf{Note 3\label{note:switch_m5}} - Tables presenting metrics on M5 contain three entries per row, i.e the WRMSSE and RMSSE metrics corresponds to the model trained using a distribution, and the MASE metric corresponds to the model trained using mean absolute error.
\end{itemize}

\subsection{Main results}

We selected TFT (Lim et al., 2021) as the main competitor to our model, because they were very close in terms of metrics and because they support both historical and future covariates, so we trained both models using the methods described above and benchmarked both using WRMSSE and MASE for M5 and WAPE and MASE for the internal dataset. We have added RMSE results on the M5 dataset, for completion. The results show Switch-Hurdle Transformer being capable of state-of-the-art performance, coming slightly ahead of the TFT models in some cases.

\begin{table}[ht]
\centering
\caption{
\textit{Benchmark metrics on the M5 dataset.}  
The table presents the performance of leading Transformer-based forecasting models under both probabilistic (WRMSSE) and point-wise (MASE) evaluation.  
The Switch-Hurdle Transformer achieves the lowest WRMSSE across all methods, demonstrating improved robustness and distributional calibration on the challenging, hierarchical M5 dataset.  
}
\begin{tabular}{|l || c | c| c|}
\hline
\textbf{Model} \rule{0pt}{1em} & \textbf{WRMSSE} & \textbf{RMSE} & \textbf{MASE}   \\
\hline
PatchTST  \rule{0pt}{1em} & 1.0393 & \textbf{2.4562} & 0.9471 \\
DeepAR             & 0.7895 & 2.9534 & 0.9087 \\
TFT                & 0.6932  & \underline{2.4686} & \textbf{0.8983}\\
TSMixer            & \underline{0.6403} & - & -\\
Switch-Hurdle (ours) & \textbf{0.6307} & 2.4744 & \underline{0.8992} \\
\hline
\end{tabular}
\label{tab:m5_top2}
\end{table}

\begin{figure*}[t]
    \centering
    \includegraphics[width=0.32\textwidth]{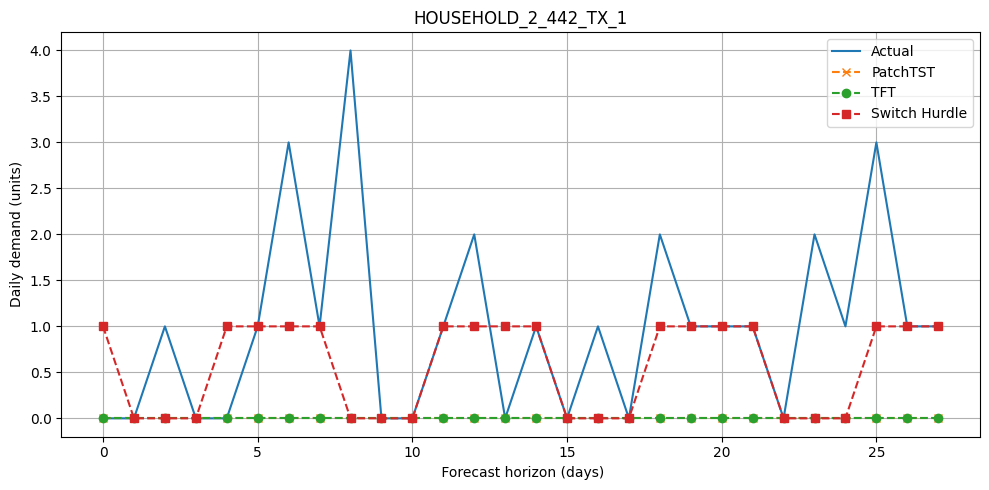}%
    \includegraphics[width=0.32\textwidth]{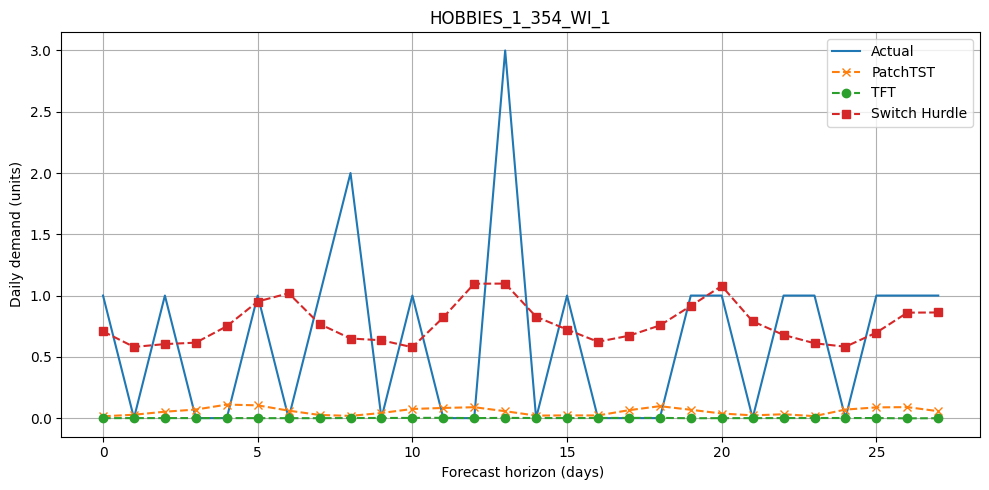}%
    \includegraphics[width=0.32\textwidth]{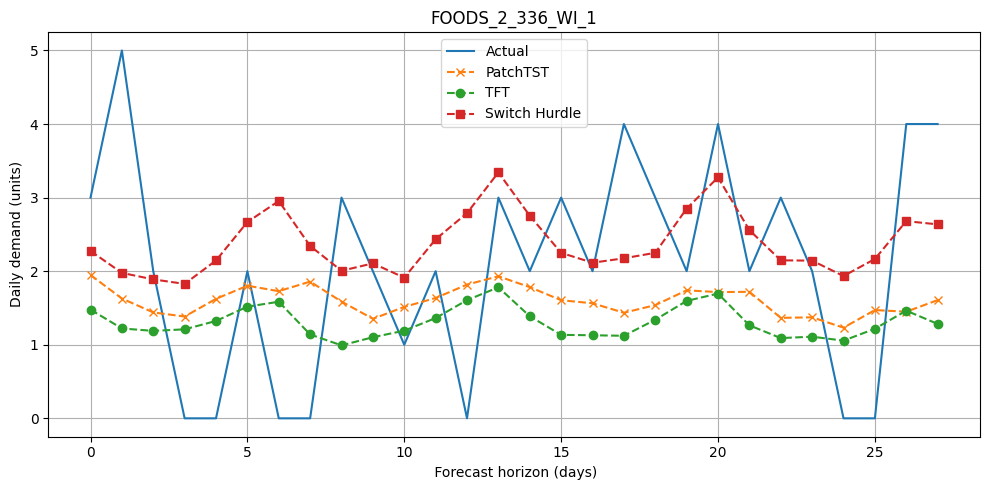}%
    \caption{
        Qualitative comparison of 28-day forecasts on three representative M5 series:
        \texttt{HOUSEHOLD\_2\_442\_TX\_1}, \texttt{HOBBIES\_1\_354\_WI\_1}, and
        \texttt{FOODS\_2\_336\_WI\_1}.
        For each series we plot the actual daily demand (blue) together with predictions from
        PatchTST, TFT, and Switch-Hurdle.
        PatchTST and TFT tend to produce nearly flat forecasts and under-react to intermittent spikes,
        while Switch-Hurdle better tracks both the occurrence and magnitude of spikes and remains close
        to zero in no-demand periods.
    }
    \label{fig:qualitative_forecasts}
\end{figure*}

\begin{table}[ht]
\centering
\caption{
\textit{Benchmark metrics on the full internal dataset.}  
This comparison highlights the model’s performance in a production-scale retail setting with rich feature space and varied demand regimes.  
The Switch-Hurdle Transformer achieves the lowest WAPE and nearly the best MASE, outperforming TFT and DeepAR while maintaining computational efficiency and stability during training.
}
\begin{tabular}{|l || c | c|}
\hline
\textbf{Model} \rule{0pt}{1em} & \textbf{WAPE} & \textbf{MASE}  \\
\hline
PatchTST \rule{0pt}{1em} & 81.22\% & 0.8478 \\
DeepAR  & 64.86\% & 0.6770 \\
TFT & \underline{55.60\%} & \textbf{0.5803} \\
Switch-Hurdle (ours) & \textbf{53.99\%} & \underline{0.5865} \\
\hline
\end{tabular}
\label{tab:internal_top}
\end{table}

\subsection{Conditional Expert Routing Analysis}
To investigate functional specialization, we analyze expert utilization \emph{conditioned} on the z-score–defined demand regime (Zero, Low, Normal, Spike). We report the conditional routing distribution
$P(e \mid \text{regime})$, normalized to 100\% per regime and aggregated over the validation split.

\begin{figure}[t]
\centering
\includegraphics[width=\linewidth]{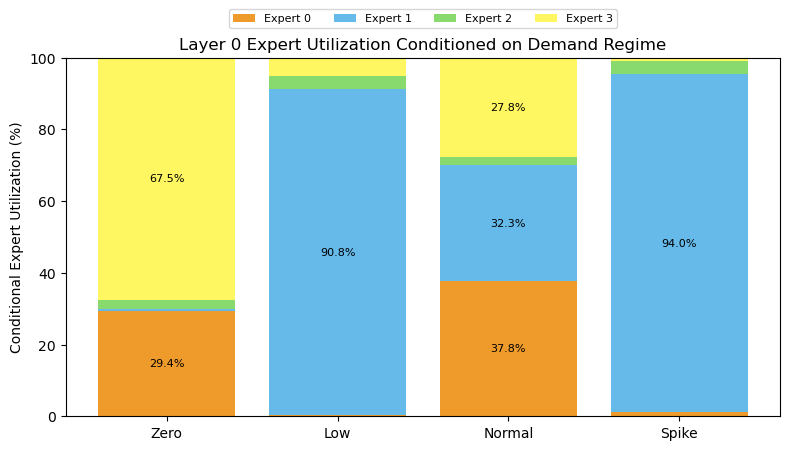}
\caption{
\textit{Layer 0 expert specialization conditioned on demand regime.}
Each bar shows $P(e \mid \text{regime})$ and is normalized to 100\% per regime (Zero, Low, Normal, Spike).
Values are normalized per regime to correct for minor measurement drift.
}
\label{fig:exp_spec_l0}
\end{figure}

As shown in Figure~\ref{fig:exp_spec_l0}, for layer 0, there is a split emerging early in the stack: Expert~3 dominates \emph{Zero} periods (67.5\%), Expert~1 captures most \emph{Low/Spike} tokens (90.8\% / 94\%), and Expert~0 contributes most in \emph{Normal} conditions (67.5\%). 
This suggests Layer~0 forms regime-aware gates that separate no-sale / rare-sale contexts from typical demand.

\begin{figure}[t]
\centering
\includegraphics[width=\linewidth]{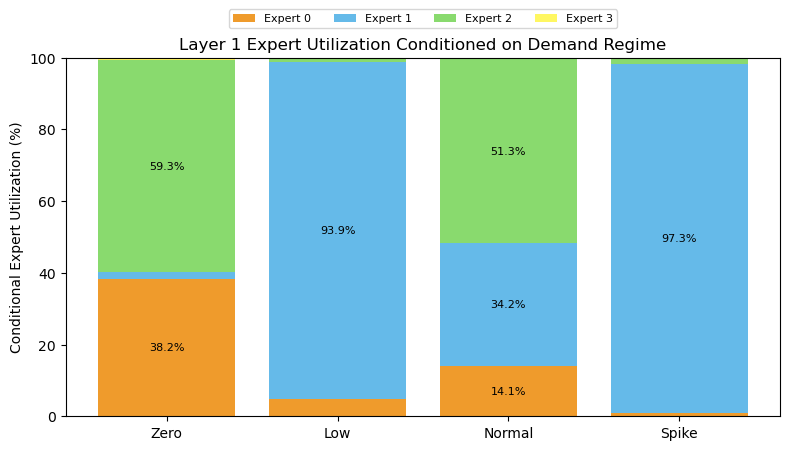}
\caption{
\textit{Layer 1 expert specialization conditioned on demand regime.}
Normalization as in Figure~\ref{fig:exp_spec_l0}.
This pattern supports the claim that Top-1 STE yields sparse forward routing with dense gradient updates,
promoting stable yet differentiated experts across layers.
}
\label{fig:exp_spec_l1}
\end{figure}

Figure~\ref{fig:exp_spec_l1} sharpens this partitioning in layer 1: Expert~2 handles the most frequent regimes, dominating \emph{Zero} (59.3\%) and \emph{Normal} (51.3\%), while Expert~1 specializes in tail events, capturing \emph{Low} (93.9\%) and \emph{Spike} (97.3\%).
Expert~3 is effectively unused across regimes, indicating stable specialization rather than collapse.

These patterns provide strong empirical evidence that Top-1 STE with KL-to-uniform regularization leads to balanced but regime-specific experts, aligning with the accuracy gains reported in the ablations (Tables~\ref{tab:sample_ablation}, \ref{tab:m5_ablation}) and main results (Tables~\ref{tab:m5_top2}, \ref{tab:internal_top}).

\subsection{Ablation studies}

To better understand the contribution of individual architectural components, we conducted a series of ablation experiments on both our internal dataset and the external M5 benchmark.
Each variant isolates a single design choice, such as expert activation function, gating mechanism, and number of experts, while keeping all the other factors constant, such as training schedule, optimizer, loss function and data splits.
This allows us to quantify the trade-offs between computational cost, convergence stability and prediction performance.

\noindent\textbf{Types of experts.} \hspace{.5em} 
The first ablation examines whether SwiGLU-based experts justify their additional computational cost compared to simpler MLP layers using GELU.
%
SwiGLUs introduce a multiplicative term that improves gradient flow and representation sparsity, helping experts to specialize in different demand regimes better.
As shown in Tables~\ref{tab:sample_ablation} and~\ref{tab:m5_ablation}, replacing SwiGLU with GELU experts slightly increases both WAPE and MASE, suggesting that simpler experts are slightly worse, especially when models are trained for longer periods.
This confirms that the non-linear gating in SwiGLU contributes meaningfully to expert diversity and effective generalization.

\noindent \textbf{Gating method.} \hspace{.5em} While our simple straight-through (ST) estimator is sparse, choosing only one expert, in the backward pass it's approximately dense because it goes through every expert, even though the results are dropped. 
%
%
This design achieves the best of both worlds: sparse expert activation for inference efficiency and dense gradient updates for stable training.
To assess the impact of this mechanism, we compared it against a soft-gating variant, where routing probabilities are used directly as continuous mixture weights.
As shown in both datasets, soft-gating results in a measurable degradation in performance: roughly +0.05 WRMSSE on M5 and +2-3\% WAPE on internal sample.
This suggests that enforcing a discrete expert selection is beneficial: it helps each expert specialize and prevents mode collapse, where all experts converge to similar behaviors.
In other words, the ST gating acts as a structural regularizer, encouraging diversity.

\noindent \textbf{Number of experts.} \hspace{.5em} 
We further examined how model capacity and depth distribution affect performance.
Our baseline uses four experts per layer across two Switch encoder layers, chosen to balance the model prediction performance and compute cost.
We then compared this baseline with a shallower configuration that varies the number of experts within a single layer:
\begin{itemize}
    \item Medium (Shallow) with 1 layer and 8 experts,
    \item Large (Shallow) with 1 layer and 32 experts.
\end{itemize}

\begin{table}[ht]
\centering
\caption{
\textit{Ablation results on the internal sample dataset.}
Each variant isolates a specific design factor in the encoder.
Results confirm that SwiGLU experts and ST Top-1 gating achieve the best trade-off between accuracy and efficiency, while excessively shallow or wide configurations degrade performance.
Lower WAPE and MASE indicate better forecasting accuracy.}
\begin{tabular}{|l || c | c|}
\hline
\textbf{Model} \rule{0pt}{1em} & \textbf{WAPE} & \textbf{MASE}  \\
\hline
Baseline \rule{0pt}{1em} & 56.97\% & 0.6184 \\
GELU Experts & 58.36\% & 0.6298 \\
Soft-gating & 59.23\% & 0.6413 \\
Medium (Shallow) & 58.02\% & 0.6298 \\
Large (Shallow) & 59.03\% & 0.6407 \\
\hline
\end{tabular}
\label{tab:sample_ablation}
\end{table}

\begin{table}[ht]
\centering
\caption{
\textit{Ablation metrics on the M5 dataset.}
Experiments on the large-scale M5 benchmark confirm the trends observed on internal data: SwiGLU experts and discrete Top-1 gating deliver better performance, while soft-gating or simpler MLP experts reduce precision.
Metrics are reported as WRMSSE (lower is better) and MASE.
}
\begin{tabular}{|l || c | c|}
\hline
\textbf{Model} \rule{0pt}{1em} & \textbf{WRMSSE} & \textbf{MASE}  \\
\hline
Baseline \rule{0pt}{1em} & 0.6307 & 0.8627 \\
GELU Experts & 0.6401 & 0.8921 \\
Soft-gating & 0.6823 & 0.9173 \\
\hline
\end{tabular}
\label{tab:m5_ablation}
\end{table}

Results in Table~\ref{tab:sample_ablation} show that scaling up the number of experts without depth degrades the model performance.
Specifically, in Large (Shallow) variant, the model performs worse than the baseline despite having 8x more experts.
We attribute this to expert underutilization: with too many experts and only one routing stage, the model struggles to learn meaningful specialization, leading to overfragmented token assignments.
Conversely, a smaller number of experts distributed across multiple layers encourage hierarchical specialization, where earlier layers capture broader demand dynamics, and later layers focus on fine-grained adjustments.
These findings highlight that depth and sparsity must be optimized simultaneously, not scaled independently.
Increasing the number of experts beyond a certain threshold offers little gain unless accompanied by deeper routing hierarchies.

The utilization patterns in Figure~\ref{fig:expert_utilization} align with the accuracy gains from hard gating (Table~\ref{tab:m5_ablation}) and SwiGLU experts (Tables~\ref{tab:sample_ablation}, \ref{tab:m5_ablation}).

The ablation findings explain better why the Switch-Hurdle Transformer consistently surpasses other architectures, including TFT, across both internal and external datasets (see Tables~\ref{tab:m5_top2} and \ref{tab:internal_top}).
By combining SwiGLU-based experts with hard Top-1 routing, the model achieves a balanced capacity allocation across demand patterns, improving its robustness to intermittent and low-signal periods that characterize retail data.
Meanwhile, the compact two-layer design ensures that computational costs remain practical for large-scale deployment.
Overall, these studies validate that the model’s performance stems not merely from increased capacity, but from architectural choices that align the biases of sparse expert specialization with the statistical structure of retail demand.

\section{Conclusion and Future Work}
\label{sec:conclusion}

This work introduced Switch-Hurdle Transformer, a novel encoder-decoder architecture for large-scale retail demand forecasting that combines dense-to-sparse expert routing with probabilistic hurdle modeling.
This model directly addresses two challenges in industrial forecasting: (i) the heavy-tailed, zero inflated type of data in demand, and (ii) the need for scalable and interpretable architectures.
Our encoder leverages a Mixture-of-Experts (MoE) design with a Top-1 routing and SwiGLU experts, enabling sparse specialization without compromising gradient stability.
The decoder's autoregressive hurdle head explicitly models zero-demand probability and positive demand magnitude through a Negative-Binomial distribution, producing probabilistic forecasts.
Together, these components deliver state-of-the-art performance on both M5 benchmark and a large proprietary dataset, outperforming established Transformer variants such as TFT and PatchTST.
Ablation studies further highlight how structured sparsity, discrete routing, and balanced depth jointly improve generalization and training efficiency.
These findings not only validate our architectural design but also shed a light on how MoE principles can be systematically applied to practical, data-sparse domains, like retail demand.

\textbf{Future Work}.
There are several promising directions for extending this work, including:
\begin{itemize}
    \item \textit{Adaptive Expert Specialization}: future research could explore dynamic expert assignment based on demand clustering, allowing the routing mechanism to evolve with market seasonality or promotions.
    \item \textit{Cross-Domain Pretraining}: similar to recent time-series foundation models, pretraining the Switch-Hurdle Transformer across multiple domains, such as retail, logistics or energy, may enhance zero-shot generalization and calibration across heterogeneous distributions.
\end{itemize}

In summary, the Switch-Hurdle Transformer demonstrates that sparse specialization and probabilistic reasoning can coexist within a single forecasting framework, yielding a model that is accurate, interpretable, and deployable at industrial scale.
We believe this work provides a foundational step toward a new generation of demand forecasting systems where neural architectures not only predict outcomes but also reveal the structure of the underlying economic processes they model.

\section*{Acknowledgments}
We thank Fabian Mu\c{s}at for implementing the initial version of Switch-Hurdle, and our colleagues in Data \& AI for helpful discussions. We also thank the reviewers for their feedback.

\bibliography{aaai2026}


\end{document}